# HYPERSPECTRAL DATA UNMIXING USING GNMF METHOD AND SPARSENESS CONSTRAINT

*Roozbeh Rajabi, Hassan Ghassemian*

ECE Department, Tarbiat Modares University, Tehran, Iran

**ABSTRACT**

Hyperspectral images contain mixed pixels due to low spatial resolution of hyperspectral sensors. Mixed pixels are pixels containing more than one distinct material called endmembers. The presence percentages of endmembers in mixed pixels are called abundance fractions. Spectral unmixing problem refers to decomposing these pixels into a set of endmembers and abundance fractions. Due to nonnegativity constraint on abundance fractions, nonnegative matrix factorization methods (NMF) have been widely used for solving spectral unmixing problem. In this paper we have used graph regularized (GNMF) method with sparseness constraint to unmix hyperspectral data. This method applied on simulated data using AVIRIS Indian Pines dataset and USGS library and results are quantified based on AAD and SAD measures. Results in comparison with other methods show that the proposed method can unmix data more effectively.

## 1. INTRODUCTION

Hyperspectral imaging has got a lot of attention in remote sensing applications. Despite high spectral resolution of hyperspectral sensors they have low spatial resolution. This is because technological restrictions in manufacturing these sensors. Low spatial resolution can cause mixed pixels to be appeared in hyperspectral images. One more reason of mixed pixels is homogeneous combination of different materials in one super pixel. First reason leads to linear mixing model and second one leads to nonlinear mixing model [1].

Mixed pixels contain more than one distinct material. These materials are called endmembers and the presence percentages of them in mixed pixels are called abundance fractions. Spectral unmixing problem aims at decomposing the measured spectra of mixed pixels into a set of endmembers and abundance fractions. Linear mixing model is often used for solving spectral unmixing problem because of its simplicity and efficiency in most cases. There are two constraints on abundance fraction values that should be considered in solving spectral unmixing problem. Abundance fraction values are always nonnegative because each endmembers can be either present in the mixed pixel (positive abundance value) or not (abundance value equals to zero). This is called nonnegativity constraint. Another one is called sum to one constraint that implies endmembers present in the mixed pixel should cover the full surface of the mixed pixel [2].

Hyperspectral unmixing has been widely addressed in the literature since the past few decades ago. There are many methods proposed for solving spectral unmixing problem. They can be categorized in geometrical, statistical and sparse regression based approaches [3]. Some methods in geometrical based category are N-FINDR [4], vertex component analysis (VCA) [5] and minimum volume simplex analysis (MVSA) [6]. Some algorithms of statistical category are Bayesian nonnegative matrix factorization with volume prior [7] and dependent component analysis (DECA) [8]. Sparse regression based methods solves spectral unmixing problem as a semi supervised problem in which spectral signature of endmembers are known in advance. Examples of this kind of algorithms are in [9] and [10].

Due to nonnegativity constraint in linear mixing model, nonnegative matrix factorization (NMF) has been widely used for solving spectral unmixing problem for example in [11] and [12]. In this paper algorithm based on graph regularized nonnegative matrix factorization [13] and sparseness constraint [12] proposed for hyperspectral data unmixing. Graph regularized NMF works based on the locality preserving assumption. Sparseness constraint uses the sparseness feature of abundance fraction values. To evaluate the proposed method simulated data is generated using AVIRIS Indian Pines dataset and USGS spectral library. Spectral unmixing results are measured based on spectral angle distance (SAD) and abundance angle distance (AAD) criteria and results compared against NMF, GNMF and NMF-SMC methods.

The paper is organized as follows. Problem is formulated in section 2. Section 3 presents the methodology framework and algorithm description of the proposed method. Experiments and results for evaluation of the algorithm using simulated data are summarized in section 4. Finally section 5 concludes the paper briefly.

## 2. PROBLEM FORMULATION

There are two main mixing models for spectral unmixing problem: Linear mixing model and Nonlinear mixing model. In linear mixing model the combination scale is macroscopic and in nonlinear model the combination scale is microscopic. Although nonlinear model can model the mixing process better, but there are obstacles to implementing this model. Therefore linear mixture model has been widely used in solving spectral unmixing problem [1].

Mathematical formulation of linear mixture model is as follows:

$$X = SA + N \quad (1)$$

Variables in (1) are summarized in Table I.

TABLE I. VARIABLES IN LINEAR MIXTURE MODEL

| Variable | Description | Dimension |
|---|---|---|
| X | Observed data | M by L |
| A | Abundance fractions | M by P |
| S | Endmember spectral signatures | P by L |
| N | Measurement noise | M by L |
| M | Total number of pixels | - |
| L | Number of spectral bands | - |
| P | Number of endmembers | - |

This model is subject to two physical constraints on abundance fraction values: nonnegativity constraint and sum to one constraint. Nonnegativity constraint implies that all abundance values should be nonnegative because an endmembers is either present in the mixed pixel (positive abundance fraction value) or not (zero abundance fraction value). Endmembers present in each mixed pixel should fully cover the surface of that pixel so sum of abundance values in each mixed pixel should equal one. This constraint is called sum to one constraint [1].

## 3. METHODOLOGY

In this section the proposed method is presented briefly. Subsection 2.1 describes original NMF method. Graph regularized NMF is presented in subsection 2.2. Sparseness measurement formulation is briefly described in subsection 2.3. Finally in 2.4 the proposed method based on previously described methods is presented.

### 2.1. Nonnegative matrix factorization (NMF)
Generally NMF problem is as follows: given a nonnegative data matrix Y, find reduced rank nonnegative matrices W and H so that [14]:

$$Y \approx WH \quad (2)$$

In spectral unmixing application W refers to spectral signature of endmembers and H refers to abundance fraction values of endmembers.

To solve this problem Euclidean distance between Y and WH can be used. Cost function based on Euclidean distance is as follows:

$$f(W, H) = \frac{1}{2} \| Y - WH \|_F^2 \quad (3)$$

Another cost function can be defined based on Kullback-Leibler divergence. Optimization on the cost function will solve the NMF problem [15].

### 2.2. Graph regularized NMF (GNMF)
Graph regularized NMF method is based on local invariance assumption. This assumption says that if two points are close in their intrinsic geometry, they will be close to each other in any other representation geometry. To model the local geometric structure, nearest neighbor graph can be used. Each vertex of the graph corresponds to a data point and is connected through an edge to its p nearest neighbors. Weight matrix on this graph can be defined using several methods. One of these methods is 0-1 weighting that means weight between two vertices equals to 1 if they are connected to each other and is 0 if they are not connected.

To consider the geometrical properties of data structure cost function using Euclidean distance can be defined as follows [13]:

$$R = \frac{1}{2} \sum_{j,l=1}^{N} \| z_j - z_l \|^2 W_{jl} \quad (4)$$

where $z_j$ and $z_l$ represents data points in the new basis.

This cost function is added to original cost function of NMF method creates the GNMF cost function. Optimization on the cost function will solve GNMF problem [13].

### 2.3. Sparseness Constraint
In [12] S-measure proposed to measure sparseness of data. S-measure is defined in the following equation:

$$S(x) = \frac{f_{max} - (k_4 - \sigma_1 k_1^2 k_2 + \sigma_2 k_1 k_3)}{f_{max} - f_{min}} \quad (5)$$

where

$$\sigma_1 > 0, \ \sigma_2 = (2\sigma_1 - 4)/3$$
$$k_1 = \|x\|_1, \ k_2 = \|x\|_2^2, \ k_3 = \|x\|_3^3, \ k_4 = \|x\|_4^4$$
$$f_{max} = \left( (1/n^3) - (\sigma_1/n) + (\sigma_2/n^2) \right) k_1^4,$$
$$f_{min} = (1 - \sigma_1 + \sigma_2) k_1^4 \quad (6)$$

In these equations $\|.\|$ denotes L-norm. S-measure takes the values between 0 and 1 smoothly. Larger S-measure corresponds to sparser vector [12].

The cost function based on S-measure is as follows:

$$J(H) = (1/n)\sum_{t=1}^{n} \text{S-measure}(h_t) \quad (7)$$

where n denotes the number of pixels in the scene.

### 2.4. Proposed Method
In this paper sparseness constraint has been applied to GNMF method and resulted cost function has been optimized to estimate the endmember and abundance fraction matrices.

The resulted cost function of proposed method is as follows:

$$f(W,H) = \frac{1}{2}\|Y - WH\|_F^2 + \alpha R - \beta J(H) \quad (8)$$

where $\alpha$ and $\beta$ are regularization parameters. $R$ and $J(H)$ are defined in equations (4) and (7) respectively.

## 4. EXPERIMENTS AND RESULTS

In this section evaluation of proposed methods is illustrated. Subsection 3.1 described the procedure of generating simulated data. Results are shown in subsection 3.2.

### 3.1. Simulated Data
For generating simulated data AVIRIS Indiana, Indian Pines dataset [16] is used. This dataset consist of 145 by 145 pixels images. Groundtruth of AVIRIS dataset (see figure 1) has been used and spectral signatures of classes has been replaced by selected spectral signatures selected (see figure 2) from USGS spectral library [17] that is available online [18]. Then images are Downsample by scale factor of 5 to generate low spatial resolution dataset and producing mixed pixels. Finally Gaussian noise with SNR=30 dB has been added to data for simulating measurement noise (see figure 3).

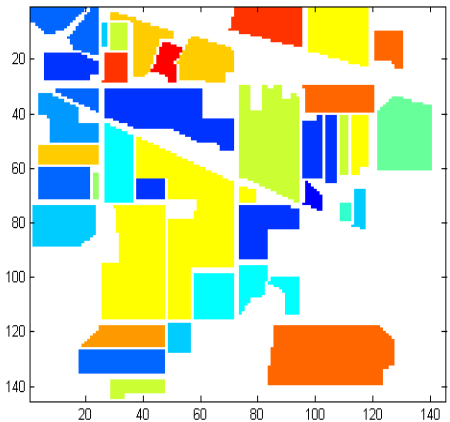

Figure 1. AVIRIS Indian Pines Groundtruth Map.

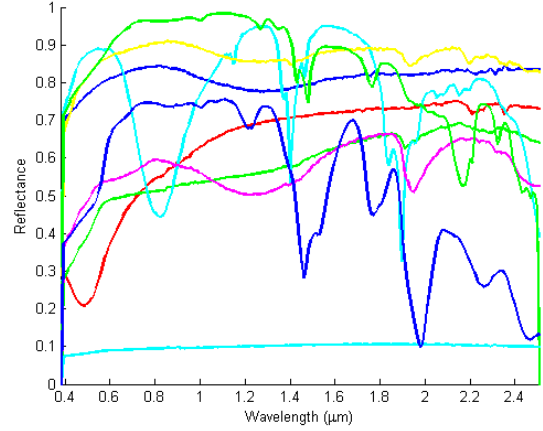

Figure 2. Selected Materials from USGS Library.

### 3.2. Results
To quantify results, two measures is used: spectral angle distance (SAD) and abundance angle distance (AAD).

$$\text{SAD}_{m_i} = \cos^{-1}\left(\frac{m_i^T \hat{m}_i}{\|m_i\|\|\hat{m}_i\|}\right) \quad (9)$$

$$\text{AAD}_{a_i} = \cos^{-1}\left(\frac{a_i^T \hat{a}_i}{\|a_i\|\|\hat{a}_i\|}\right) \quad (10)$$

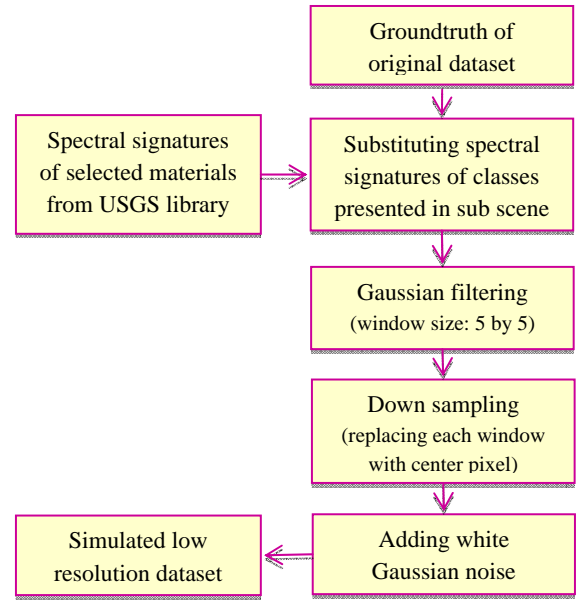

Figure 3. Algorithm for generating simulated data

$m_i$ represents *i*th endmember spectral signature and $a_i$ represents *i*th pixel abundance fractions. These measures are calculated for each endmember and each pixel. RMS values should be calculated to obtain overall measures.

$$\text{rms}_{\text{SAD}} = \left( \frac{1}{p} \sum_{i=1}^{p} (\text{SAD}_{m_i})^2 \right)^{1/2} \quad (11)$$

$$\text{rms}_{\text{AAD}} = \left( \frac{1}{N} \sum_{i=1}^{N} (\text{AAD}_{a_i})^2 \right)^{1/2} \quad (12)$$

where p is the number of endmembers and N is the number of pixels in the scene.

Table I summarizes the comparison results based on defined criteria for NMF-SMC and the proposed method (GNMF-SMC). Results show that GNMF-SMC outperforms other results used in this evaluation.

TABLE II. COMPARISON RESULTS BASED ON RMS VALUES IN DEGREES

|  | SAD | AAD |
|---|---|---|
| NMF | 19.54 | 10.23 |
| GNMF | 15.76 | 8.34 |
| NMF-SMC | 13.24 | 6.67 |
| GNMF-SMC | 11.87 | 5.89 |

## 5. CONCLUSION

Mixed pixels in hyperspectral images consist of more than one endmembers. Spectral unmixing refers to decomposing mixed pixels spectra into a set of endmembers and abundance fractions. In this paper new method for spectral unmixing of hyperspectral data has been proposed. This method works based on graph regularized NMF (GNMF) and sparseness constraint. The proposed method has been applied on simulated dataset generated using AVIRIS dataset and USGS spectral library. Results are presented in terms of SAD and AAD measures and compared against some other methods. Comparison results show that the proposed algorithm can effectively unmix hyperspectral data. Future work includes evaluation of the algorithm on real dataset, implementation of the cost function using Kullback-Liebler divergence and using other weighting function in neighborhood graph.